\documentclass[11pt,a4paper]{article}

\usepackage[acceptedWithA]{tacl2018v2}

\usepackage{times}
\usepackage{latexsym}

\usepackage[linesnumbered,ruled,vlined]{algorithm2e}

\usepackage{microtype}

\usepackage{booktabs}
\usepackage{soul}
\usepackage{tabularx}
\usepackage{bbm}
\usepackage{threeparttable}
\usepackage{amsmath}
\usepackage{amssymb}
\usepackage{graphicx}
\usepackage[normalem]{ulem}
\useunder{\uline}{\ul}{}

\newcolumntype{Y}{>{\centering\arraybackslash}X}

\newcommand{\secref}[1]{\S\ref{#1}}
\newcommand{\sys}{ColBERT}
\newcommand{\sysqa}{ColBERT-QA}

\renewcommand{\textit}[1]{\emph{#1}}

\newcommand{\tablesubsectiondiv}{
    \specialrule{\heavyrulewidth}{4pt}{\defaultaddspace}}

\newcommand{\tablesubsectionheadrule}{\midrule \addlinespace[0.5ex]}

\newcommand{\tablemodeldiv}{\midrule[0.1pt]}

\title{Relevance-guided Supervision for OpenQA with ColBERT} %

\author{Omar Khattab \\
  Stanford University \\
  \scalebox{.9}{\texttt{okhattab@stanford.edu}} \\\And
  Christopher Potts \\
  Stanford University \\
  \scalebox{.9}{\texttt{cgpotts@stanford.edu}} \\\And
  Matei Zaharia \\
  Stanford University \\
  \scalebox{.9}{\texttt{matei@cs.stanford.edu}} \\}

\date{}

\usepackage{array}
\newcolumntype{!}{>{\global\let\currentrowstyle\relax}}
\newcolumntype{^}{>{\currentrowstyle}}
\newcommand{\rowstyle}[1]{\gdef\currentrowstyle{#1}%
  #1\ignorespaces
}

\begin{document}
\maketitle
\global\csname @topnum\endcsname 0
\global\csname @botnum\endcsname 0

\begin{abstract}
Systems for Open-Domain Question Answering (OpenQA) generally depend on a \emph{retriever} for finding candidate passages in a large corpus and a \emph{reader} for extracting answers from those passages. In much recent work, the retriever is a learned component that uses coarse-grained vector representations of questions and passages. We argue that this modeling choice is insufficiently expressive for dealing with the complexity of natural language questions.
To address this, we define \sysqa, which adapts the scalable neural retrieval model \sys\ to OpenQA. \sys\ creates fine-grained interactions between questions and passages.
We propose an efficient weak supervision strategy that iteratively uses \sys{} to create its own training data. This greatly improves OpenQA retrieval on Natural Questions, SQuAD, and TriviaQA, and the resulting system attains state-of-the-art extractive OpenQA performance on all three datasets.
\end{abstract}

\section{Introduction}
\label{sec:introduction}

\begin{figure}[t]
\centering
\includegraphics[width=\columnwidth]{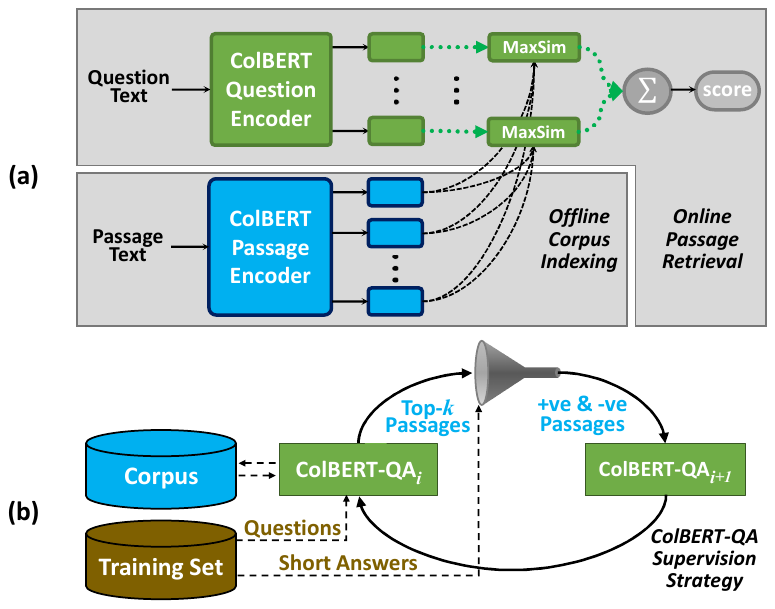}
\vspace*{-3mm}
\caption{
Sub-figure (a) depicts the \sys{} retrieval model \cite{khattab2020colbert}. \sys{} encodes questions and passages into multiple embeddings and allows those to interact via a scalable maximum-similarity (MaxSim) mechanism.
Sub-figure (b) illustrates our proposed \sysqa{} training strategy: we use an existing retrieval model to collect the top-$k$ passages for every training question and, with a simple heuristic, sort these passages into positive (+ve) and negative (--ve) examples, using those to train another, more effective retriever. This process is applied thrice, and the resulting \sysqa{} is used in the OpenQA pipeline.}
\label{fig:pipeline}
\vspace{-3mm}
\end{figure}

The goal of Open-Domain Question Answering (OpenQA; \citealt{Voorhees:Tice:2000}) is to find answers to factoid questions in potentially massive unstructured text corpora. Systems for OpenQA typically depend on two major components: a \emph{retrieval model} to find passages that are relevant to the user's question and a \emph{machine reading model} to try to find an answer to the question in the retrieved passages. At its best, this should combine the power and scalability of current information retrieval (IR) models with recent advances in machine reading comprehension (MRC). However, if the notions of relevance embedded in the IR model fail to align with the requirements of question answering, the MRC model will not reliably see the best passages and the system will perform poorly.

Many prior approaches to OpenQA rely on classical IR models (e.g., BM25;~\citealt{robertson1995okapi}) whose notions of relevance are not tailored to questions. In effect, this reduces the OpenQA problem to few-passage MRC, imposing a hard limit on the quality of the passages seen by the MRC model. Recent work has sought to address this problem by \emph{learning} to retrieve passages. For instance, \citet{guu2020realm} and \citet{karpukhin2020dense} jointly train vector representations of both passages and questions to support similarity-based retrieval. This has led to state-of-the-art performance on multiple OpenQA datasets.

However, existing OpenQA retrievers exhibit two major limitations. First, the representations learned by these models are relatively coarse: they encode each passage into a \emph{single} high-dimensional vector and estimate relevance via one dot-product. We argue this is not expressive enough for reliably matching complex natural-language questions to their answers. Second, existing systems present substantial tradeoffs when it comes to supervision: they expect hand-labeled ``positive'' passages, which may not always exist; or they use simple models like BM25 to sample ``positives'' and ``negatives'' for training, which may provide weak positives and unchallenging negatives; or they conduct retrieval within the training loop, which requires frequently re-indexing a large corpus (e.g., tens or hundreds of times in REALM; \citealt{guu2020realm}) or a frozen document encoder that cannot adapt to the task (e.g., as in RAG; \citealt{lewis2020retrieval}).  We argue that supervision methods in this space need to be more flexible and more scalable.

We tackle both limitations with \sysqa.\footnote{\scalebox{.8}{\url{https://github.com/stanfordnlp/ColBERT-QA}}} To address the first problem, we leverage the recent neural retrieval model \sys\ \citep{khattab2020colbert} to create an end-to-end system for OpenQA. Like other recent neural IR models, \sys\ encodes both the question and the document using BERT \citep{devlin-etal-2019-bert}. However, a defining characteristic of \sys\ is that it explicitly models \emph{fine-grained interactions} (Figure~\ref{fig:pipeline}(a)) between the question and document representations (\secref{sec:method}), in particular by comparing each question term embedding with each passage term embedding. Crucially, \sys{} does so while scaling to millions of documents and maintaining low query latency. We hypothesize that this form of interaction will permit our model to be sensitive to the nature of questions without compromising the OpenQA goal of scaling to massive datasets.

To address the second problem, we propose \textit{relevance-guided supervision} (RGS), an efficient weak-supervision strategy that allows the retriever to guide its own training \textit{without} frequent re-indexing or freezing the document encoder. Instead of expensive pretraining, RGS starts from an existing weak retrieval model (e.g., BM25) to collect the top-$k$ passages for every training question and uses a provided weak heuristic to sort these passages into positive and negative examples, relying on the ordering imposed by the retriever. These examples are then used to train a more effective retriever, and this process is applied 2--3 times, with the resulting retriever deployed in the OpenQA pipeline. Crucially for scaling to large corpora, RGS only requires re-indexing the corpus once or twice during training and correspondingly only retrieves positives and negatives in 2--3 large batches. In doing so, RGS permits fine-tuning of the document encoder during all of training, freeing it to co-adapt with the query encoder to the task's complexities.

To assess \sysqa, we report on experiments with Natural Questions \citep{kwiatkowski-etal-2019-natural}, SQuAD \citep{rajpurkar-etal-2016-squad}, and TriviaQA \citep{joshi-etal-2017-triviaqa}. We adopt the OpenQA formulation in which the passage is not given directly as gold evidence, but rather must be retrieved, including during training. Further, we focus on the standard \textit{extractive} OpenQA setup, where the reader model extracts the answer string from one of the retrieved passages. On all three datasets, \sysqa\ achieves state-of-the-art retrieval and extractive OpenQA results.

In summary, our contributions are:

\begin{enumerate}
    \item We propose relevance-guided supervision (RGS), an efficient iterative strategy for fine-tuning a retriever given a weak heuristic.

    \item We conduct the first systematic comparison between ColBERT's fine-grained interaction and recent single-vector retrievers like DPR. We find that ColBERT exhibits strong transfer learning performance to new OpenQA datasets and that fine-tuned ColBERT delivers large gains over existing OpenQA retrievers.
    
    \item We apply RGS to \sys{} and a single-vector retriever, and find that each improves by up to 2.3 and 3.2 points in Success@20, respectively. Our resulting \sysqa{} system establishes state-of-the-art retrieval and downstream performance.
    
\end{enumerate}

\begin{figure}[t]
\centering
\includegraphics[width=\columnwidth]{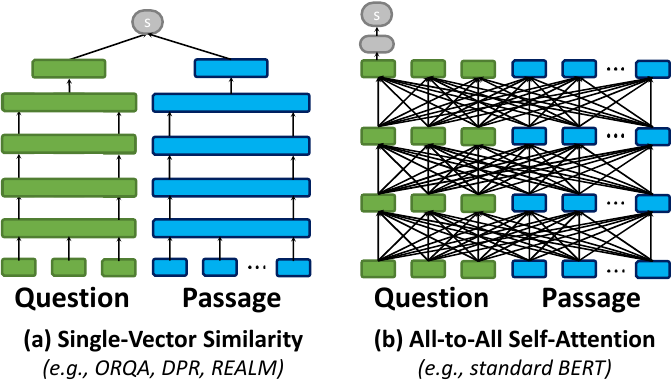}
\vspace*{-5mm}
\caption{A comparison between two extremes of building neural retrievers with Transformers. On the left, single-vector models independently encode each question and passage into a vector and model relevance as one dot-product. On the right, all-to-all attention models feed a sequence concatenating the question and each passage through the encoder, using layers of self-attention to estimate a relevance score. Diagrams adapted from \citet{khattab2020colbert} with permission.}
\label{fig:comparison}
\vspace{-3mm}
\end{figure}

\section{Background \& Related Work}
\label{sec:background}

\subsection{Machine Reading Comprehension}
\label{sec:background:MRC}

MRC refers to a family of tasks that involve answering questions about about text passages \citep{Clark_Etzioni_2016}. In recent work, the potential answers are usually selected from a set of pre-defined options \citep{hirschman-etal-1999-deep,richardson-etal-2013-mctest,iyyer-etal-2014-neural} or guaranteed to be a substring of the passage \citep{yang-etal-2015-wikiqa,rajpurkar-etal-2016-squad,kwiatkowski-etal-2019-natural,joshi-etal-2017-triviaqa}. Here, we start from the version of the task established by \citeauthor{yang-etal-2015-wikiqa} and \citeauthor{rajpurkar-etal-2016-squad}, in which the passage $p$ and question $q$ are MRC inputs and the correct answer $\hat{a}$ is a literal substring of $p$.

\subsection{OpenQA}
\label{sec:background:mainstream-OpenQA}

In its most general formulation, the OpenQA task is defined as follows: given a large corpus $C$ of documents (e.g., Wikipedia or a fraction of the Web) and a question $q$, the goal is to produce the correct short answer $\hat{a}$. Much of the earliest work in question answering adopted this paradigm \citep{Voorhees:Tice:2000,Watson:2010,kushman-etal-2014-learning}. Our focus is on the specific version of OpenQA that is a relaxation of the standard MRC paradigm \citep{chen2017reading,kratzwald-feuerriegel-2018-adaptive}: the passage is no longer given, but rather needs to be retrieved from a corpus, and the MRC component must seek answers in the retrieved passages without a guarantee that an answer is present in any of them.

\subsection{Retrieval Models for OpenQA}
\label{sec:background:OpenQA-retrieval}

Mainstream approaches use sparse retrieval models (e.g., BM25; \citealt{robertson2009probabilistic}) and various heuristics (e.g., traversal of Wikipedia hyperlinks) to retrieve a set of $k$ passages that are relevant to the question $q$. This (closed) set of passages is often \textit{re-ranked} with a Transformer encoder \citep{Vaswani-etal:2017} to maximize precision, as in \citet{wang2019multi} and \citet{yang2019end}. Subsequently, mainstream approaches deploy a reader to read each of these $k$ passages and return the highest-scoring answer, $\hat{a}$, present as a span in one or more of the $k$ passages. This approach is represented in \citet{chen2017reading}, \citet{min2019discrete}, and \citet{asai2019learning}, among many others.

More recent literature has shown that there is value in learning the retriever, including ORQA~\cite{lee2019latent}, REALM~\cite{guu2020realm}, and DPR~\cite{karpukhin2020dense}. Common to all three is the BERT two-tower architecture depicted in Figure~\ref{fig:comparison}(a). This architecture encodes every question $q$ or passage $d$ into a single, high-dimensional vector and models relevance via a single dot-product. However, these models differ substantively in supervision.

\citet{lee2019latent} propose the Inverse Close Task (ICT), a self-supervised task to pretrain these encoders for retrieval. The question encoder is fed a random sentence $q$ from the corpus and the passage encoder is fed a number of context passages, one of which is the true context of $q$, and the model learns to identify the correct context. \citet{guu2020realm} extend this task to retrieval-augmented language modeling (REALM), where the encoders are optimized to retrieve passages that help with a Masked Language Modeling task \citep{devlin-etal-2019-bert}. After pretraining, both ORQA and REALM freeze the passages index and encoder, and fine-tune the question encoder to retrieve passages that help a jointly-trained reader model extract the correct answer. Though self-supervised, the pretraining procedures of ORQA and especially REALM are computationally expensive, owing to the amount of data they must see and as REALM re-indexes the corpus during pretraining every 500 steps (out of 200k). %

Very recently, \citet{karpukhin2020dense} propose a dense passage retriever (DPR) that directly trains the architecture in Figure~\ref{fig:comparison}(a) for retrieval, relying on a simple approach to collect positives and negatives. For every question $q$ in the training set, \citet{karpukhin2020dense} recover the hand-labeled evidence passages (if available; otherwise they use BM25-based weak supervision) as positive passages and sample negatives from the top-$k$ results by BM25. Importantly, they also use in-batch negatives: every positive passage in the training batch is a negative for all but its own question. Using this simple strategy, DPR considerably outperforms both ORQA and REALM and established a new state-of-the-art for extractive OpenQA.

The single-vector approach taken by ORQA, REALM, and DPR can be very fast but allows for only shallow interactions between $q$ and $d$. \citet{khattab2020colbert} describe a spectrum of options for encoding and scoring in neural retrieval, within which this single-vector paradigm is one extreme. At the other end of this spectrum is the model shown in Figure~\ref{fig:comparison}(b). Namely, we could feed BERT a concatenated sequence $\langle q, d \rangle$ for every passage $d$ to be ranked and fine-tune all BERT parameters against the ranking objective. This allows for very rich interactions between the query and document. However, such a model is prohibitively expensive beyond \textit{re-ranking} a small set of passages already retrieved by much cheaper models. 

\begin{figure}[t]
\centering
\includegraphics[width=0.8\columnwidth]{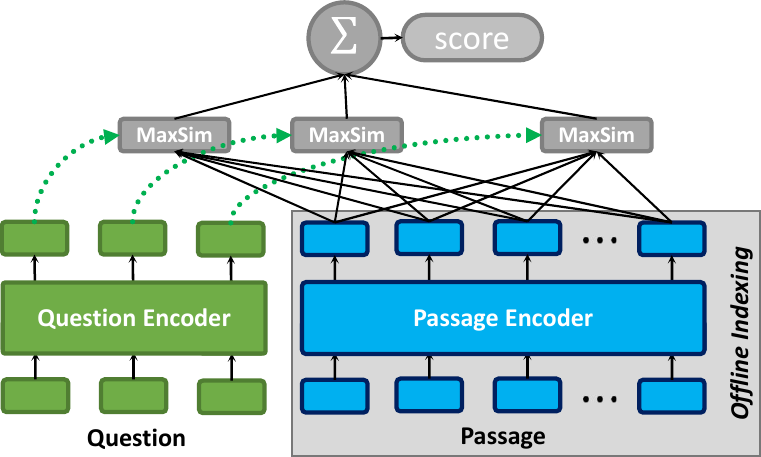}
\vspace*{-3mm}
\caption{The general architecture of \sys{} given a question and a passage. Diagram adapted from \citet{khattab2020colbert} with permission.}
\label{fig:ColBERT}
\vspace{-1mm}
\end{figure}

\sys\ (Figure~\ref{fig:ColBERT}) seeks a middle ground: it separately encodes the query and document into token-level representations and relies on a scalable scoring mechanism that creates rich interactions between $q$ and $d$. Central to their efficiency, the interactions are \emph{late} in that they involve just the output states of BERT. Essential to their quality, they are \emph{fine-grained} in that they cross-match token representations of $q$ and $d$ against each other.

Our work differs from \citet{khattab2020colbert} in two major ways. First, they assume gold-evidence positives, which may not exist in OpenQA, and use BM25 negatives, which we argue is insufficient for an end-to-end retriever. We propose a simple and efficient strategy, namely RGS, that improves training quality. Second, \citet{khattab2020colbert} report gains against a single-vector ablation of their IR system, but we ask if these gains hold against (concurrent) state-of-the-art single-vector models in OpenQA, where a reader could in principle overshadow retrieval gains. Our work confirms that late interaction is superior even (if not especially) in OpenQA: we report considerable gains when using traditional supervision and even larger gains with RGS. We also report strong results when using an out-of-domain transfer learning setting from IR, which work by \citet{akkalyoncu-yilmaz-etal-2019-cross} considers but in the context of neural re-rankers and between standard IR tasks.

\subsection{Supervision Paradigms in OpenQA}
\label{sec:background:OpenQA-supervision}

Broadly, there are two paradigms for training and evaluating OpenQA models.

\textbf{Gold-Evidence Supervision.} Some OpenQA datasets supply annotated \textit{evidence passages} or human-curated contexts from which the gold answers can be derived. For such datasets, it is natural to rely on these labels to train the retriever, akin to typical supervision in many IR tasks. For example, Natural Questions contains labeled ``long answers'', which DPR uses as positive passages. 

However, using gold-evidence labels is not possible with OpenQA datasets that only supply question--answer string pairs, like TriviaQA and WebQuestions \cite{berant-etal-2013-semantic}.\footnote{TriviaQA provides automatic ``distant supervision'' labels.} Moreover, manual annotations might fail to reflect the richness of passages answering the same question in the corpus, possibly due to biases in how passages are found and annotated. 

\textbf{Weak Supervision.} Addressing these limitations, weakly-supervised OpenQA \cite{lee2019latent,guu2020realm} supplies its own evidence passages during training. To do so, it exploits a question's short answer string as a crucial supervision signal. For a question $q$ in the training set, a passage that contains $q$'s answer string $\hat{a}$ is treated as a potential candidate for a positive passage. To weed out \emph{spurious} matches, additional strategies are often introduced. We categorize those into \textit{retrieve-and-filter} and \textit{inner-loop retrieval} strategies. In retrieve-and-filter, an existing retriever and a weak heuristic are essentially intersected to find promising passages for training. For instance, \citet{wang2019multi} and \citet{karpukhin2020dense} consider only passages highly-ranked by BM25 for $q$ as candidates, where the answer string can be a more reliable signal.

In inner-loop retrieval, the training loop retrieves passages for each example using the model being fine-tuned. This is conducted by ORQA, REALM, and RAG~\cite{lewis2020retrieval}, approaching ``end-to-end'' training of the retriever. However, such inner-loop retrieval requires major approximations, since it is infeasible to compute forward and backward passes over the entire collection for every training batch. Here, ORQA, REALM, and RAG \textit{freeze} their document encoder (and the indexed vectors) when fine-tuning for OpenQA, which restricts the adaptability of the model to this task and/or to new corpora. During pretraining, REALM does not freeze the document encoder, but then it has to very frequently re-index the corpus with training, and the method suffers quality loss if the index is allowed to become stale. 

While existing retrieve-and-filter approaches reflect the naive term-matching biases of BM25, and existing inner-loop retrieval strategies restrict training the document encoder or require frequent re-indexing and repeated retrieval, RGS offers a scalable and effective alternative that allows the retriever itself to collect the training examples while fine-tuning the document encoder and only re-indexing 1--2 additional times.

\label{sec:background:semi-supervision}

Weak supervision is also a topic in IR. For instance, \citet{dehghani2017neural} explore using BM25 as a teacher model to train a neural ranker. Other weak signals in IR include anchor text~\cite{zhang2020selective} and headings text \cite{macavaney2019content}, treated as queries for which the target content is assumed relevant. Lastly, user interactions have long been used for supervision in search and recommendation. However, the gold answer string is unique to OpenQA and we show it can lead to large quality gains when combined with an effective retriever.

\section{\sysqa{}}
\label{sec:method}

We now describe our \sysqa{} system. We propose \textit{relevance-guided supervision} (RGS) (\secref{sec:method:supervision}), a scalable strategy that uses the retriever being trained and a weak heuristic to gather training examples in a few discrete rounds. As Figure~\ref{fig:pipeline} illustrates, we use RGS to fine-tune \sys{} models in three stages arriving at \textbf{\sysqa$_1$}, \textbf{\sysqa$_2$}, and \textbf{\sysqa$_3$}.

\subsection{The \sys{} Model}
\label{sec:method:modeling}

\sys{} capitalizes on BERT's capacity for contextually encoding text into token-level output representations. Given a question $q$, we tokenize it as $[q_{1}, \ldots, q_{n}]$. The token sequence is truncated if it is longer than $N$ (e.g., $N=32$) or padded with \texttt{[MASK]} tokens if it is shorter. \citet{khattab2020colbert} refer to this padding as \textit{query augmentation} and show that it improves \sys{}'s effectiveness. These tokens are processed by BERT to obtain a sequence of output vectors $\mathbf{q} = [\mathbf{q}_{1}, \ldots, \mathbf{q}_{N}]$.

The encoding of a passage $d$ given tokens $[d_{1}, \ldots, d_{m}]$ follows the same pattern but no augmentation is performed. BERT processes these into output vectors $\mathbf{d} = [\mathbf{d}_{1}, \ldots, \mathbf{d}_{m}]$. For both $\mathbf{q}$ and $\mathbf{d}$, we apply a linear layer on top of each representation to control the output dimension. Each vector is finally rescaled into unit length. 

Let $E_q$ (length $N$) and $E_d$ (length $m$) be the final vector sequences derived from $\mathbf{q}$ and $\mathbf{d}$. The \sys\ scoring mechanism is given as follows:
\begin{equation}
\label{eq:scorer}
S_{q,d} = \sum_{i=1}^{N} \max_{j=1}^{m} E_{q_i} \cdot E_{d_j}^{T}
\end{equation}

As illustrated in Figure~\ref{fig:ColBERT}, for every query embedding, \sys{} computes its maximum-similarity (MaxSim) score over all embeddings in the passage, then sums these scores. This mechanism ``softly'' matches each query term with a passage term, producing a similarity score. Crucially, this computation scales to billions of tokens corresponding to many millions of passages, permitting \sys{} to retrieve passages directly from a massive corpus, not restricted by the limitations of a ``first-stage'' retriever (e.g., BM25 or single-vector models).

Overall, \sys{} balances the scalability of single-vector retrievers and the richness of BERT's query--document interaction (\secref{sec:background:OpenQA-retrieval}). For IR, \citet{khattab2020colbert} show that \sys{} outperforms a single-vector retriever and that, while \sys{}'s precision is comparable to that of BERT when re-ranking a closed set of passages (e.g., top-1000 passages retrieved by BM25), its scalability (allowing full-corpus retrieval) boosts recall.

\sys{} is trained to give positive passages higher scores than negative passages. More specifically, it requires triples $\langle q, d^+, d^- \rangle$, where $q$ is a short query, $d^+$ is a positive passage, and $d^-$ is a negative passage. We score each of $d^+$ and $d^-$ according to Equation~\ref{eq:scorer} given $q$, and treat this as a binary classification task, optimizing the model parameters using a cross-entropy loss. For generic IR applications, negative passages can be sampled uniformly from the top-$k$ (e.g., $k = 1000$) results for an existing IR system. Positive passages are more difficult to obtain; IR evaluations generally require \textit{labeled} positives as provided by datasets like MS MARCO Ranking \citep{nguyen2016ms}.

\subsection{Relevance-Guided Supervision} 
\label{sec:method:supervision}

\begin{algorithm}
\small
\DontPrintSemicolon
\SetAlgoLined
\SetNoFillComment

\KwIn{Training questions $Q$ and corpus $C$}
\KwIn{Supervision heuristic $H$}
\KwIn{Retrieval model $R_0$, number of rounds $n$}
\KwOut{Stronger retriever $R_n$} \vspace{2mm}

 \For{\textnormal{round} $t \gets 1$ \textbf{to} $n$}{

 Index corpus $C$ for retrieval with $R_t$. \;
 
 $Rank_t \gets \{ R_i.retrieve(q) : q \in Q \}$ \;
 
 $P_t = \{H.getPositives(r[0 : k^+]) [0:t]$ \\  \hspace{30pt}  $: r \in Rank_t \}$ \;
 
 $N_t = \{H.getNegatives(r[0 : k^-])$ \\  \hspace{30pt}  $: r \in Rank_t\}$ \;
 
 $T_t = \{ \langle q, p, n \rangle : q \in Q,$ \\ \hspace{30pt} $p \in P_t[q], n \in N_t[q] \} $ \;
 
 Train a new retriever $R_{t+1}$ using triples $T_t$.
 
 }

 \Return{Final retriever $R_n$}\;
 
 \caption{Relevance-guided Supervision, given a weak heuristic $H$}
 \label{algo:RGS}
\end{algorithm}

OpenQA often lacks gold-labeled passages and instead has \textit{short answer strings} as supervision signals. This signal is known to lead to many spurious matches and existing work tackles this by either BM25-based retrieve-and-filter, which can be simplistic, or inner-loop retrieval methods, which can be expensive and restrictive (\secref{sec:background:OpenQA-supervision}).

Algorithm~\ref{algo:RGS} outlines the proposed RGS, which seeks to alleviate these limitations by \textit{outer-loop} retrieval. RGS takes as input a set of training question $Q$, a corpus of passages $C$, and an initial weak retrieval model (e.g., BM25). In general, RGS assumes a task-specific weak heuristic $H$, which in this work indicates whether a passage contains a question's short-answer string.  RGS proceeds in $n$ discrete rounds (we set $n=3$ in this work). Each round uses as a building block the retrieve-and-filter mechanism. For every question $q$ in the \textit{training} set, we retrieve $R$'s top-$k$ results (Line 3). Then, we take as positives (Line 4) the highest-ranked $t$ (e.g., $t=3$) passages \textit{that contain $q$'s short answer}, restricted to the top-$k^+$ (e.g., $k^+=20$) results. Every passage in the top-$k^-$ (potentially with $k^- \gg k^+$) that does not contain the short answer is treated as a negative (Line 6). We train a new retriever with these examples (Line 10), and repeat until all rounds are complete.

Our key hypothesis is that more effective retrieval would collect more accurate and diverse positives and more challenging and realistic negatives. Importantly, RGS collects positives and negatives entirely outside the training loop. As a result, we can flexibly sample positives and negatives from large depths and we can fine-tune the entire model (i.e., including the document encoder) without having to frequently re-index the corpus. However, applying this process more than once on the training set risks overfitting in a way that rewards a narrow set of positives and drifts in the negatives it samples. To avoid this, we use a deterministic split of the training set such that no model sees the same question during both training and positive/negative retrieval. To this end, we create arbitrary equal-sized splits of all training sets before starting RGS and train each \sysqa{} model with 50\% of the training questions.\footnote{We note that, if desired, it is easy to extend this procedure to train a \sysqa{} model on all questions: train two independent copies of the supervising model $R$, one for each half of the training set, and combine their rankings to supervise the final model. We leave such improvements to future work.}

To bootstrap \sysqa{}, \secref{sec:eval-retriever} selects $R_0$ as the bag-of-words model BM25. Training yields \sysqa$_1$, which enables us to establish \sys{}'s high quality even with simple weak supervision. We then apply RGS in two rounds, leading to \sysqa$_2$ and \sysqa$_3$, whose retrieval quality consistently outperforms \sysqa$_1$ and ultimately surpasses existing retrievers by large margins.

\subsection{Reader Supervision for OpenQA}
\label{sec:method:reader-supervision}

Like many OpenQA systems, \sysqa\ uses a BERT encoder as a reader. After retrieval, the reader takes as input a concatenation of a question $q$ and a passage $d$: \texttt{[CLS]} $q$ \texttt{[SEP]} $d$ \texttt{[SEP]}, for each passage $d$ in the top-$k$ set retrieved. The reader scores each individual short span $s$ in each passage $d$. Similar to \citet{lee2019latent}, we model the probability of each span $s$, namely $P(s | d)$, as $P(s | d) \propto \texttt{MLP}(h_{\texttt{start}(s)};h_{\texttt{end}(s)})$ for the output embeddings at the start and end tokens of span $s$.

To train the reader, we use a set of triples $\langle q, d^+, d^- \rangle$ for every question $q$ in the training set with positive $d^+$ and negative $d^-$. We collect these triples using the same general heuristic used for retrieval training, extracting them from the ranking model $R$ whose passages we feed to the reader during inference. We treat every span matching the answer in $d^+$ as correct for optimization and minimize the maximum marginal likelihood of the correct answer, normalizing the span probabilities over the spans in both documents. For the subset of spans $\hat{S}$ that match the gold answer in $d^+$, we minimize the loss $-\log \sum_{s \in \hat{S}} P(s | d^+)$.

\begin{table*}[t]
\small
\centering
\begin{threeparttable}
\begin{tabularx}{\textwidth}{@{}!l^c^X^X^X^X^X^X@{}} 
\toprule
\rowstyle{\bfseries}

    \textbf{Name}    &
    \begin{tabular}{@{}c@{}}\textbf{Source(s) of} \\ \textbf{Positives / Negatives}\end{tabular} &
    
    \multicolumn{3}{c}{
        \begin{tabular}{@{}c^c^c@{}}
            \multicolumn{3}{c}{ \textbf{Success@20 (test)} } \\
            \textbf{NQ} & TQ & SQ \\
        \end{tabular}
    }
    &
    \multicolumn{3}{c}{
        \begin{tabular}{@{}c^c^c@{}}
            \multicolumn{3}{c}{ \textbf{EM (dev)} } \\
            \textbf{NQ} & TQ & SQ \\
        \end{tabular}
    }
\\

\tablesubsectiondiv
\multicolumn{8}{c}{
\textbf{Baseline Retrievers}
} \\
\tablesubsectionheadrule

    BM25 (Anserini) & n/a &
    64.0 & 77.3 & 71.4 &
    - & - & - \\
    
    DPR \scalebox{.85}{(core variant) \cite{karpukhin2020dense}} & Gold / BM25 &
    78.4 & 79.4 & 63.2 &
    - & - & - \\
     
    DPR \scalebox{.85}{(best variant) \cite{karpukhin2020dense}} & Gold / BM25 &
    79.4 & 79.9 & 71.5 &
    - & - & - \\
    
    Single-vector \sys{} \scalebox{.85}{(ablation)} & BM25 &
    78.3 & 82.8 & 72.9 &
    - & - & -  \\

\tablesubsectiondiv
\multicolumn{8}{c}{
\textbf{\sys{}-guided Supervision Retrievers (ours)}
} \\
\tablesubsectionheadrule

    Out-of-Domain \sysqa{} & MS MARCO &
    79.1 & 80.3 & 76.5 &
    - & - & - \\
    
    \sysqa{}$_1$ & BM25 &
    82.9 & 84.7 & 82.1 &
    47.1 & 69.5 & 49.4 \\ 
    
    \sysqa{}$_2$ & \sysqa{}$_1$ &
    85.0$^\ddagger$ & 85.3$^\ddagger$ & \textbf{83.9}$^\ddagger$  &
    \textbf{48.6} & \textbf{70.2} & 51.0 \\ 
    
    \sysqa{}$_3$ & \sysqa{}$_2$ &
    \textbf{85.3}$^\ddagger$ & \textbf{85.6}$^\ddagger$ & 83.7$^\ddagger$ &
    47.9 & 69.8 & \textbf{51.8} \\ 

\bottomrule
\end{tabularx}

\caption{A comparison between baseline retrieval models and approaches for training \sysqa{}.
A subscript $1$, $2$, or $3$ on \sysqa{} indicates the number of fine-tuning rounds for OpenQA retrieval. We report Success@20 on the test sets for comparison with \citet{karpukhin2020dense}'s results. We reserve the test-set EM evaluation for \secref{sec:eval-reader}. For Success@20, $\ddagger$ indicates significant improvement over \sysqa{}$_1$; details in the main text.}
\label{tab:retrieval}

\end{threeparttable}
\end{table*}

\begin{table}[]
\centering
\resizebox{\columnwidth}{!}{%
\begin{tabular}{@{}lclclcl@{}}
\toprule
\textbf{Metric} &
  \multicolumn{2}{c}{\textbf{NQ}} &
  \multicolumn{2}{c}{\textbf{TQ}} &
  \multicolumn{2}{c}{\textbf{SQ}} \\
\textbf{} &
  \textbf{C3} &
  \multicolumn{1}{c}{\textbf{$\Delta_{C1}$}}&
  \textbf{C3} &
  \multicolumn{1}{c}{\textbf{$\Delta_{C1}$}} &
  \textbf{C3} &
  \multicolumn{1}{c}{\textbf{$\Delta_{C1}$}} \\ \midrule

$k$=1  &  52.9  &  +5.7  &  68.0  &  +2.6  &  56.0  &  +2.1  \\
$k$=5  &  75.6  &  +4.2  &  80.7  &  +1.6  &  75.1  &  +2.1  \\
$k$=10  &  81.4  &  +2.8  &  83.6  &  +1.3  &  79.6  &  +1.4  \\
$k$=20  &  85.3  &  +2.3  &  85.6  &  +0.9  &  83.7  &  +1.6  \\
$k$=50  &  88.6  &  +1.6  &  87.4  &  +0.5  &  87.7  &  +1.8  \\
$k$=100  &  90.1  &  +1.1  &  88.4  &  +0.2  &  89.4  &  +1.4  \\ \midrule
MRR@100  &  62.8  &  +4.8  &  73.7  &  +2.2  &  64.5  &  +2.1  \\

\bottomrule
\end{tabular}%
}
\caption{Test-set retrieval quality of \sysqa{}$_3$ (with gains over \sysqa{}$_1$) at various depths.}
\label{tab:retrieval.depths}

\vspace{-3mm}

\end{table}

\section{Evaluating \sysqa{}'s Retrieval}
\label{sec:eval-retriever}

\sysqa\ presents many options for training the retriever so it extracts the best passages for the reader. This section explores a range of these options to address the following key questions.

Our first question concerns \sys{}'s retrieval modeling capacity. Can \sys{}'s fine-grained interactions improve the accuracy of OpenQA retrieval when tuned for this task? We consider this under a weak supervision paradigm: training based on BM25 ranking. In \secref{sec:evaluation:baselines}, we find that \sys\ is highly effective in this setting, considerably outperforming classical and single-vector retrieval.

Our second question concerns an out-of-domain version of our model. \sys\ is standardly trained to perform IR tasks. Is this form of transfer learning from IR sufficient for effective OpenQA? If so, then this might be an appealingly modular option for many applications, allowing system designers to focus on the reader. In \secref{sec:evaluation:zero-shot}, we show that \sys\ succeeds in this setting.

Our third set of questions concerns how best to supervise \sysqa{} for OpenQA. In particular, can relevance-guided supervision improve on standard BM25-based supervision by capitalizing on the structure and effectiveness of \sys{}? We find that the answer is ``yes''; \sysqa{} with RGS consistently proves to be the best method (\secref{sec:evaluation:colbert-guided-supervision}), while only requiring 1--2 additional iterations of indexing and training.

\subsection{Methods}

Similar to \citet{chen2017reading}, \citet{guu2020realm}, and \citet{karpukhin2020dense}, we study the retrieval quality in OpenQA by reporting Success@$k$ (S@$k$ for short), namely, the percentage of questions for which a retrieved passage (up to depth $k$) contains the gold answer string. This metric reflects an upper bound on the performance of an extractive reader that reads $k$ passages, assuming it always locates the answer if present. However, readers in practice are affected by the quality and number of passages that contain the answer string and by passages that do \textit{not} contain the answer. To evaluate this more directly, we suggest employing a BERT-large model (with ``whole-word-masking'' pretraining) as a reader, trained for each retriever as described in~\secref{sec:method:reader-supervision}. We report the exact-match (EM) score corresponding to each retriever--reader combination on the validation set. Hyperparmeter tuning is described in~\secref{app:details}.

We conduct these experiments on the open versions of the Natural Questions (NQ), SQuAD (SQ), and TriviaQA (TQ) datasets. Additional details on these datasets are in our appendices. Our retrieval results are summarized in Table~\ref{tab:retrieval}. To describe how each model was trained, we report its sources of positive and negative passages. 

\subsection{Baselines}
\label{sec:evaluation:baselines}

The top of Table~\ref{tab:retrieval} shows the retrieval quality of our baselines: BM25 and the recent state-of-the-art DPR. As \citet{karpukhin2020dense} report, DPR has a considerable edge over BM25 on NQ and TQ but not SQuAD. The authors attribute the weak performance on SQuAD to the presence of high lexical matching between its questions and evidence passages. Indeed, as \citet{lee2019latent} argues, this might stem from the SQuAD annotators writing the questions after reading the passage. %

Unlike single-vector models, \sys{} is capable of fine-grained contextual matching between the words in the question $q$ and the passage $d$ so, by design, it can \textit{softly} match contextual representations of words across $q$ and $d$. The results on NQ, TQ, and SQ confirm that fine-grained modeling exceeds the quality attained with single-vector representations, consistently outperforming the baselines across all three datasets.

As described in~\secref{sec:background:OpenQA-retrieval}, DPR is trained using in-batch negatives and differs from \sys{} in a number of ways (e.g., weak vs.~gold supervision on NQ, the depth of negatives used in training, etc.). To better account for differences, we include a ``single-vector \sys{}'' ablation trained in the same manner as our \sysqa{}$_1$ model. This ablation is similar to the ablation model evaluated by \cite{khattab2020colbert} on MS MARCO and it resembles DPR in that it uses a dot-product of a 768-dimensional vector extracted from BERT's \texttt{[CLS]} token for each query or passage. Unlike DPR, we do not use in-batch negatives for this model to more directly compare with \sys{}.\footnote{Unlike our \sys{} models, we found this single-vector ablation performs better under \textit{re-ranking} BM25 top-1000 than under full-corpus retrieval on the validation sets; hence, we report its re-ranking performance.} The results validate that fine-grained interaction is a stronger retrieval choice for OpenQA.

\subsection{Evaluating \sys{} out of domain}
\label{sec:evaluation:zero-shot}

We now directly examine out-of-domain \sysqa{} in Table~\ref{tab:retrieval}. For this, we train \sys{} as in \citet{khattab2020colbert} on the MS MARCO Ranking dataset, a natural choice for transfer learning to OpenQA. To elaborate, it is originally derived from the MRC dataset with the same name and contains many question-like search queries as well as other ``free-form'' queries. Unless otherwise stated, we evaluate \sys{}-based models by indexing and retrieving from the full corpus.

The results in Table~\ref{tab:retrieval} show that out-of-domain \sysqa{} outperforms BM25 and, in fact, is already competitive with the DPR retriever across the three datasets.\footnote{If we plug \sysqa{}$_1$'s retriever over this out-of-domain retrieval, we get 45.8, 66.5, and 44.8 EM (dev) on NQ, TQ, and SQ, respectively.} To further contextualize these results, we note that out-of-domain \sysqa{} is more effective than ``zero-shot'' ORQA and REALM as reported in \cite{guu2020realm}: its development-set Success@5 exceeds 65\% on NQ whereas both zero-shot ORQA and REALM are below 40\%, on a different random train/dev split of the same data. We conclude that transfer learning with models having strong inductive biases, like \sys{}'s matching, offers a promising alternative to expensive unsupervised retrieval pretraining.

\subsection{Evaluating Relevance-Guided Supervision}
\label{sec:evaluation:colbert-guided-supervision}

We now turn our attention to the proposed RGS strategy, defined in~\secref{sec:method:supervision}. RGS seeks to exploit \sys{}'s high precision to collect high quality positives and challenging negative and to leverage its recall to collect richer positives with a wider coverage of questions. Below, we test how these hypotheses fare after one, two, and three stages of fine-tuning. We first focus on Success@20 then consider the EM metric. We start the fine-tuning of each \sys{} retriever from BERT-base.

We bootstrap \sysqa{} in a weakly-supervised fashion on each target dataset: \textbf{\sysqa{}$_1$} relies on BM25 for collecting its positive and negative passages. As reported in Table~\ref{tab:retrieval}, it already outperforms every baseline considered so far, including DPR, by considerable margins.  This strong performance emphasizes that \sys{}'s capacity to learn from \textit{noisy} supervision examples can lead to more accurate OpenQA retrieval. It also suggests that adapting to the characteristics of OpenQA-oriented retrieval, the corpus, and the target QA datasets can beat high-quality retrieval training for generic IR. This is evidenced by the consistent gains of \sysqa{}$_1$ over the already-effective out-of-domain model.

Next, we apply RGS by using \sysqa$_{1}$ itself to create data for further training, yielding \textbf{\sysqa$_2$}, and take this a step further in using \sysqa$_{2}$ to supervise \textbf{\sysqa$_{3}$}. As shown, \sysqa$_2$ and \sysqa$_3$ are consistently the most effective, while only requiring that we index and retrieve positives and negatives one or two times. In particular, they raise test-set Success@20 over the next best, \sysqa{}$_1$, by up to 0.9--2.4 points on the three datasets, a sizeable increase in the fraction of questions that can be answered by an extractive reader. We conduct a more granular comparison between \sysqa$_{3}$ and \sysqa$_{1}$ in Table~\secref{tab:retrieval.depths}, which shows that the gains are even larger at smaller depths (e.g., up to 5.7 points in S@1), a property useful in applications where a reader only has the budget to consume a few passages per question.\footnote{With the Wilcoxon signed-rank test (with $p < 0.05$ and Bonferroni correction) on Success@20, we find that \sysqa{}$_2$ and \sysqa{}$_3$ are always statistically significantly better than \sysqa{}$_1$ (pre-correction p-values all less than 0.0001) and detect no such difference between \sysqa{}$_2$ and \sysqa{}$_3$.}

As mentioned in~\secref{sec:method}, relying on \sys{}'s scalability permits us to leverage its accuracy by applying it to the entire corpus, enabling improved recall. For instance, if we use \sysqa{}$_3$ for re-ranking BM25 top-1000 instead of end-to-end, full-corpus retrieval, its Success@20 score drops on the development set from 84.3\% to just 80.7\%.

It is standard to evaluate OpenQA retrievers using the occurrence of answers in the retrieved passages. Having done this, we consider another evaluation paradigm: the \textit{development-set EM scores} resulting from training a reader corresponding to each retriever. The EM scores in Table~\secref{tab:retrieval} reinforce the value of RGS (i.e., \sysqa{}$_2$ and \sysqa$_3$ outperforming \sysqa$_1$ by up to 0.7--2.4 EM points). Simultaneously, the EM scores show that a higher fraction of answerable questions (i.e., higher Success@$k$) does not always imply higher EM with a particular reader. We hypothesize that this depends on the correlation of the retriever and reader mistakes: a stronger retriever may find more ``positive'' passages but it can also retrieve more challenging negatives that could confuse an imperfect reader model. Considering this, we next evaluate the full \sysqa{} system against existing end-to-end OpenQA systems.

\begin{table*}[t]
\small
\centering
\begin{threeparttable}

\begin{tabularx}{0.98\textwidth}{@{}!X^c^c^c^r^r^r@{}} 
\toprule
\rowstyle{\bfseries}
Name    &
\begin{tabular}{@{}c@{}}Retriever \\ \scalebox{.8}{(\# parameters)}\end{tabular} &
\begin{tabular}{@{}c@{}}Reader \\ \scalebox{.8}{(\# parameters)}\end{tabular}  &
\begin{tabular}{@{}c@{}}NQ\end{tabular} &
\begin{tabular}{@{}c@{}}TQ\end{tabular} &
\begin{tabular}{@{}c@{}}SQuAD\end{tabular} & \\
\tablesubsectiondiv
\multicolumn{6}{c}{\textbf{Mainstream OpenQA}}                \\
\tablesubsectionheadrule
\begin{tabular}{@{}l@{}}BM25 + BERT   \\  \citet{karpukhin2020dense} \end{tabular}                    & 
BM25 &
\begin{tabular}{@{}c@{}}BERT-base \\ \scalebox{.8}{(110M)}\end{tabular}
& 32.6  & 52.4  & 38.1 / -
\\

\tablemodeldiv
\begin{tabular}{@{}l@{}}GraphRetriever   \\  \citet{min2019knowledge} \end{tabular}                    & 
\begin{tabular}{@{}c@{}}BM25 \\ +graph-based retr.\end{tabular} &
\begin{tabular}{@{}c@{}}BERT-base \\ \scalebox{.8}{(110M)}\end{tabular}
& 34.5 &  56.0 & -  \\

\tablemodeldiv
\begin{tabular}{@{}l@{}}PathRetriever   \\  \citet{asai2019learning} \end{tabular}                    & 
\begin{tabular}{@{}c@{}}TFIDF \\ +graph-based retr.\end{tabular} &
\begin{tabular}{@{}c@{}}BERT-WWM \\ \scalebox{.8}{(330M)}\end{tabular}
& 32.6  & - & - / 56.5 \\
\tablesubsectiondiv
\multicolumn{6}{c}{\textbf{Learned-Retrieval OpenQA}}                \\
\tablesubsectionheadrule
\begin{tabular}{@{}l@{}}ORQA   \\  \citet{lee2019latent} \end{tabular}                    & 
\begin{tabular}{@{}c@{}}ORQA \\ \scalebox{.8}{(2$\times$110M)}\end{tabular} &
\begin{tabular}{@{}c@{}}BERT-base \\ \scalebox{.8}{(110M)}\end{tabular}
&  33.3  & 45.0 & 20.2 / -  \\
\tablemodeldiv
\begin{tabular}{@{}l@{}}REALM   \\  \citet{guu2020realm} \end{tabular}                    & 
\begin{tabular}{@{}c@{}}REALM \\ \scalebox{.8}{(2$\times$110M)}\end{tabular} &
\begin{tabular}{@{}c@{}}BERT-base \\ \scalebox{.8}{(110M)}\end{tabular}
&  40.4 & - & - \\
\tablemodeldiv
\begin{tabular}{@{}l@{}}DPR (best variant)  \\  \citet{karpukhin2020dense} \end{tabular}                    & 
\begin{tabular}{@{}c@{}}DPR \\ \scalebox{.8}{(2$\times$110M)}\end{tabular} &
\begin{tabular}{@{}c@{}}BERT-base \\ \scalebox{.8}{(110M)}\end{tabular}
&  41.5 & 57.9 & 36.7 / -  \\
\tablemodeldiv
\begin{tabular}{@{}l@{}}RAG (generative)   \\  \citet{lewis2020retrieval} \end{tabular}                    & 
\begin{tabular}{@{}c@{}}DPR (joint) \\ \scalebox{.8}{(2$\times$110M)}\end{tabular} &
\begin{tabular}{@{}c@{}}BART-large \\ \scalebox{.8}{(406M)}\end{tabular}
& 44.5   & 56.1 & - \\
\tablesubsectiondiv

\multicolumn{6}{c}{\textbf{Our Models}}                \\
\tablesubsectionheadrule

\sysqa{} (base) &
\begin{tabular}{@{}c@{}}\sysqa{}$_3$ \\ \scalebox{.8}{(110M)}\end{tabular} &
\begin{tabular}{@{}c@{}}BERT-base \\ \scalebox{.8}{(110M)}\end{tabular}
& \underline{42.3}  & \underline{64.6} & \underline{47.7} / \underline{53.5} \\
\tablemodeldiv
\sysqa{} (large) &
\begin{tabular}{@{}c@{}}\sysqa{}$_3$ \\ \scalebox{.8}{(110M)}\end{tabular} &
\begin{tabular}{@{}c@{}}BERT-WWM \\ \scalebox{.8}{(330M)}\end{tabular} &
\textbf{47.8}  & \textbf{70.1} & \textbf{54.7} / \textbf{58.7} & \\
\bottomrule
\end{tabularx}

\caption{End-to-end OpenQA results, reporting exact-match (EM) on the open-domain test sets. For SQuAD, we report EM on the main 2018 Wikipedia dump (left) and on the 2016 Wikipedia dump (right). Underlined are the best results using only a BERT-base reader; bold is overall best.}
\label{tab:evaluation:reader}

\end{threeparttable}

\vspace{-3mm}
\end{table*}

\section{End-to-end OpenQA Evaluation}
\label{sec:eval-reader}

Our primary question is at this point is whether \sysqa{}'s retrieval improvements lead to better end-to-end OpenQA quality. We find that the answer is consistently positive: \sysqa\ leads to state-of-the-art extractive OpenQA results.

\subsection{Datasets}

We continue to evaluate on NQ, SQuAD, and TQ, using their \textit{open} versions \cite{lee2019latent}, which discard the evidence passages of the validation and test questions. As is standard, for each of the three datasets, we randomly split the original training set for training and validation and use the original development set for testing. 

Like the results in~\secref{sec:eval-retriever}, for all three datasets, we primarily conduct our evaluation on the December 2018 Wikipedia dump used by the majority of our baselines. However, \citet{asai2019learning} argue that for SQuAD, comparisons should be made using another common 2016 dump closer to the creation date of this dataset. Thus, for SQuAD, Table~\ref{tab:evaluation:reader} additionally includes our end-to-end results on this 2016 Wikipedia dump (details in~\secref{app:datasets}).

\subsection{Baselines}

The results are shown in Table~\ref{tab:evaluation:reader}. We report Exact Match (EM) of the gold answer on the test sets of the OpenQA versions of NQ, TQ, and SQuAD. We compare against both mainstream OpenQA approaches that rely on heuristic retrieval and recent systems that use learned vector-based retrieval.

Under mainstream OpenQA, we first report the purely BM25-retrieval-based results of \citet{karpukhin2020dense}. Next, we report GraphRetriever \cite{min2019knowledge} and PathRetriever \cite{asai2019learning}, both of which propose graph-based augmentation mechanisms to classical bag-of-words IR models (in particular, BM25 and TF-IDF, respectively). Unlike most other OpenQA systems included, PathRetriver uses a BERT-large reader (pretrained with whole-word masking or WWM) with 330M parameters.

Under learned-retrieval OpenQA, we report results of ORQA, REALM, and DPR, which are extractive OpenQA models like \sysqa{}. We also include the concurrent work RAG by \citet{lewis2020retrieval}: unlike all other models included here, RAG is a \textit{generative} model that uses a BART-large reader (406M parameters; \citealt{lewis2019bart}).

\subsection{Our Models}

We use \sysqa$_{3}$ as our retriever and test two reader models: (a) \textbf{\sysqa{} (base)} with a BERT-base reader and (b) \textbf{\sysqa{} (large)} with a BERT-large reader pretrained with whole-word masking (WWM). The former allows us to compare directly with approaches that use BERT-base. The latter allows us to evaluate \sysqa{}'s performance given a powerful reader that can better leverage its retrieval and permits comparing against the recent RAG model, which uses over 626M parameters, and PathRetriever, which also uses BERT-WWM. We train our readers separately from the retrievers, but use the retriever to collect triples for reader supervision.

As Table~\ref{tab:evaluation:reader} shows, \sysqa{} (base) outperforms the existing models that use a BERT-base reader on all three datasets. Moreover, it also outperforms the generative RAG on TQ, despite using far fewer parameters and being restricted to extractive reading. In fact, \sysqa{} (base) already attains state-of-the-art extractive OpenQA performance on TQ and SQuAD.

Looking at \sysqa{} (large) next, we observe that it also outperforms all of our baselines, including RAG, and establishes state-of-the-art performance for extractive OpenQA on all three datasets. These results reinforce the impact of improved, weakly-supervised retrieval as explored in~\secref{sec:eval-retriever}. Importantly, \sysqa{}'s gains are consistent and substantial: every baseline---besides REALM, which does not evaluate on SQuAD or TQ---trails \sysqa{} by several EM points on at least one dataset.

\section{Conclusion}
\label{sec:conclusion}

We proposed \sysqa, an end-to-end system for OpenQA that employs the recent \sys{} retrieval model to improve both retriever modeling and supervision in OpenQA. To this end, we developed a simple yet effective training strategy that progressively improves the quality of the OpenQA training data, even without hand-labeled evidence passages. Our results show that \sys{} is highly effective for OpenQA retrieval; with the proposed training paradigm, \sysqa{} can improve retrieval quality over existing methods by over five Success@20 points and the resulting OpenQA pipeline attains state-of-the-art extractive OpenQA results across three popular datasets.

\begin{footnotesize}
\textbf{Acknowledgments:}  We would like to thank Ashwin Paranjape for valuable discussions and feedback. This research was supported in part by affiliate members and other supporters of the Stanford DAWN project---Ant Financial, Facebook, Google, Infosys, NEC, and VMware---as well as Cisco, SAP, a Stanford HAI Cloud Credit grant from AWS, and the NSF under CAREER grant CNS-1651570. Any opinions, findings, and conclusions or recommendations expressed in this material are those of the authors and do not necessarily reflect the views of the National Science Foundation.
\end{footnotesize}

\appendix

\begin{table*}[]
\centering
\resizebox{\textwidth}{!}{%
\begin{tabular}{@{}llccccl@{}}
\toprule
\textbf{} &
  \multicolumn{1}{c}{\textbf{}} &
  \multicolumn{1}{c}{\textbf{S@1}} &
  \multicolumn{3}{c}{\textbf{Delta of C3 over}} &
  \multicolumn{1}{c}{\textbf{}} \\ \cmidrule(lr){3-6}
\textbf{Slice} &
  \multicolumn{1}{l}{\textbf{Size}} &
  \multicolumn{1}{c}{\textbf{C3}} &
  \multicolumn{1}{c}{\textbf{C1}} &
  \multicolumn{1}{c}{\textbf{DPR}} &
  \multicolumn{1}{c}{\textbf{BM25}} &
  \multicolumn{1}{l}{\textbf{Example (where C3 outperforms all baselines)}} \\ \midrule
all  &  8757  &  53  &  +5  &  +8  &  +30  &  \footnotesize{who sang i won't give up on you}  \\

\midrule

misc.  &  1144  &  45  &  +2  &  +7  &  +21  &  \footnotesize{poems that use the first letter of a word}  \\
what  &  1116  &  47  &  +5  &  +9  &  +25  &  \footnotesize{what 's the major league baseball record for games won in a row}  \\
who  &  3651  &  59  &  +6  &  +9  &  +33  &  \footnotesize{who wrote the song ruby sung by kenny rogers}  \\
where  &  712  &  53  &  +4  &  +5  &  +33  &  \footnotesize{where does the united states keep an emergency stockpile of oil quizlet}  \\
when  &  1818  &  50  &  +7  &  +6  &  +34  &  \footnotesize{when did wales last win the 6 nations}  \\
superlative  &  628  &  52  &  +7  &  +20  &  +35  &  \footnotesize{who formed the highest social class of republican and early imperial rome}  \\

 \bottomrule
\end{tabular}%
}
\caption{Retrieval results by ``slice'' on the NQ validation set. For each slice, the table reports the number of queries (Size), Success@1 on NQ of ColBERT-QA3 (C3), and the improvement of this model over three baselines: ColBERT-QA1 (C1), DPR, and BM25. Each row contains an example query, for which C3 gets a correct passage and all three baselines fail. Rows are sorted by delta over BM25.}
\label{tab:analysis:slicing}

\vspace{-3mm}

\end{table*}

\section*{Appendices}

\section{Datasets}\label{app:datasets}

\textbf{Natural Questions} (NQ; \citealt{kwiatkowski-etal-2019-natural}) contains real questions submitted to Google by multiple searchers each, filtered such that a Wikipedia article is among the top-5 results. Answers are short spans in the Wikipedia article. Owing to its large scale and organic nature, this is the main dataset in our experiments. 

\textbf{SQuAD v1.1}~\cite{rajpurkar2016squad} is the popular QA dataset whose questions were written by crowdworkers over Wikipedia articles.

\textbf{TriviaQA} (TQ; \citealt{joshi2017triviaqa}) is a set of trivia questions originally created by enthusiasts.

Each of the three datasets has approx. 79k training questions and 9k validation questions; NQ has 4k test questions and both of SQuAD and TQ have 11k test questions. For all three datasets, we use the train/validation splits released by \citet{karpukhin2020dense}. We note that the \textit{test sets} are the same for all methods across both datasets.

\section{Implementation Details}\label{app:details}

\textbf{Corpus \& Preprocessing.} Similar to related work, we use the full English Wikipedia, excluding tables, lists, and disambiguation pages. To facilitate comparisons with the state of the art, we use the preprocessed passages released by \citet{karpukhin2020dense} for the 20 December 2018 dump.\footnote{\scalebox{.7}{\url{https://github.com/facebookresearch/DPR}}} We use this corpus unless otherwise stated. For the experiments that need Wikipedia 2016, we preprocess the 2016 dump released by \citet{chen2017reading} in the same manner.

Following standard practice (e.g., as in \citealt{karpukhin2020dense} and \citealt{lee2019latent}), we prepend the title of each Wikipedia page to all of its passages. For retrieval evaluation, we treat a passage as relevant if it contains the short answer string one or more times. Unlike ours, \citet{karpukhin2020dense}'s open-source implementation does not take into account the title when evaluating whether a passage contains the answer or not. We found that this has a minimal impact on Success@$k$ for various depths $k$. For instance, while \sysqa{}$_3$ outperforms the best DPR results reported by the authors by 5.7--12.2 points in S@20, this delta would be reduced by only 0.1--0.6 points if we did not consider answer matches in passage titles.

\textbf{Hyperparameters.} For BM25 retrieval, we use the Anserini~\cite{yang2018anserini} toolkit with its default MSMARCO-tuned $k_1$ and $b$ for passage search. We train our models using Python 3 and PyTorch 1.6, relying on the popular HuggingFace transformers library for the pretrained BERT models \citep{wolf-etal-2020-transformers}. We apply PyTorch's built-in automatic mixed precision for faster training and inference. For training \sys{} models, we use a batch size of 64 triples (i.e., a question, a positive passage, and a negative passage each) with the default learning rate of $3\times10^{-6}$ and default dimension $d=128$ for each embedding. Our implementation starts from the ColBERT code.\footnote{\scalebox{.7}{\url{https://github.com/stanford-futuredata/ColBERT/}}}

For our retrievers, we select the best checkpoint among $\{10k, 20k, ..., 50k\}$ via Success@20 on a sample of 1500 questions from each validation set. For \sysqa{}$_2$ and \sysqa{}$_3$, the last checkpoint was consistently the best-performing, suggesting more training could lead to even higher accuracy. For RGS, we set the depth $k^-$ to 1000 passages (for sampling negatives) and sample the highest-ranked $t=5$ positives from the top $k^+=50$ passages. If no positives exist in the top-$k^+$, we take the highest-ranked positive from the top-1000. For training the readers, we set $t =3 $ and $k^+ = k^- = 30$. For training our reader models, we use batches of 32 triples with a learning rate of $1\times10^{-5}$. For the base and large reader, we select the best checkpoint among $\{10k, 20k, ..., 50k\}$ and $\{10k, 20k\}$, respectively, via EM on the validation set, and tune the number of passages fed to the reader during inference in $\{5, 10, 15, 20, 30, 50\}$.

\section{Retrieval Analysis}

Across a number of simple ``slices''~\cite{chen2019slice} of NQ-dev, Table~\ref{tab:analysis:slicing} compares \sysqa{}$_3$ (C3 for short) against \sysqa{}$_1$ (C1), DPR, and BM25, revealing gains due to RGS (vs.~C1), ColBERT (vs.~DPR), and neural retrieval (vs.~BM25). We use the Success@1 metric, stressing each model's precision. We find that C3's gains are stable and consistent across all models and slices, with gains of 2--7 points against C1, 7--20 points against DPR, and 21--35 against BM25. Interestingly, the largest gains against all three can be seen on the \textit{superlative} slice, which contains over 600 queries that have comparative markers such as ``best'', ``most'', or ``-est'' (e.g., ``largest'').

\section{RGS for Single-Vector Retrieval}
\begin{table}[]
\centering
\begin{tabularx}{\columnwidth}{@{}Xccc@{}}
\toprule
\textbf{Ablation Name}        & \multicolumn{3}{c}{\textbf{Success@20 (test)}}                                           \\
\textbf{}            & \textbf{NQ}          & \multicolumn{1}{c}{\textbf{TQ}} & \multicolumn{1}{c}{\textbf{SQ}} \\ \midrule

SV Round 1 (end-to-end)  &  74.0  &  78.1  &  59.2   \\
SV Round 2 (end-to-end) &  76.6  &  78.4  &  \textbf{61.7}   \\
SV Round 3 (end-to-end)  &  \textbf{77.2}  &  \textbf{79.4}  &  61.5   \\

\bottomrule
\end{tabularx}%
\caption{The results of applying RGS to a \textit{single-vector} ablation of the \sys{} model.}
\label{tab:retrieval.SV}
\end{table}

In this experiment, we apply RGS to a \textit{single-vector} (SV) ablation of \sys{}. Table~\secref{tab:retrieval.SV} shows the results, extending those in Table~\ref{tab:retrieval}. Like DPR but unlike our ablation in Table~\ref{tab:retrieval}, we test with end-to-end retrieval (and not by re-ranking BM25) and use in-batch negatives (but unlike DPR, only collect those on a per-GPU basis in a four-GPU setting). We observe very similar gain patterns to applying RGS over \sys{}, with +3.2, +1.3, and +2.3 point gains between the first and third rounds. The absolute results are considerably weaker than \sysqa{}'s, pointing to the value of late interaction, and are on average 1.0-point weaker than DPR's core setting, suggesting better tuning or implementation of single-vector modeling.

\section{Computational Cost \& Latency}

We conducted our experiments using servers with four Titan V (12GB) or 4--8 V100 (16/32GB) GPUs. We report running times for using four Titan V GPUs with our latest implementation. Each round of retriever training and validation requires 7--8 hours. Precomputing passage representations and indexing into FAISS~\cite{johnson2019billion} requires approximately 6 hours. We apply our retrieval in batch mode: retrieval with all NQ train/dev/test questions takes 2--3 hours. Python scripts for pre- and post-processing (e.g., labeling as positive/negative, creating triples) add up to a few hours but leave much room for optimizations.

We also compare the (one-query) retrieval latency of \sys{} against our single-vector ablation, noting our Python-based research implementations are not optimized for production. Both models use the same underlying testbed, which treats single-vector representations as a special case of \sys{}. For questions in NQ-dev, retrieval latency is 71ms and 36ms per query for \sys{} and for the single-vector ablation, respectively.

\newpage
\clearpage

\bibliography{anthology,bibliography}

\begin{thebibliography}{40}
\expandafter\ifx\csname natexlab\endcsname\relax\def\natexlab#1{#1}\fi

\bibitem[{Akkalyoncu~Yilmaz et~al.(2019)Akkalyoncu~Yilmaz, Yang, Zhang, and
  Lin}]{akkalyoncu-yilmaz-etal-2019-cross}
Zeynep Akkalyoncu~Yilmaz, Wei Yang, Haotian Zhang, and Jimmy Lin. 2019.
\newblock \href {https://doi.org/10.18653/v1/D19-1352} {Cross-domain modeling
  of sentence-level evidence for document retrieval}.
\newblock In \emph{Proceedings of the 2019 Conference on Empirical Methods in
  Natural Language Processing and the 9th International Joint Conference on
  Natural Language Processing (EMNLP-IJCNLP)}, pages 3490--3496, Hong Kong,
  China. Association for Computational Linguistics.

\bibitem[{Asai et~al.(2020)Asai, Hashimoto, Hajishirzi, Socher, and
  Xiong}]{asai2019learning}
Akari Asai, Kazuma Hashimoto, Hannaneh Hajishirzi, Richard Socher, and Caiming
  Xiong. 2020.
\newblock Learning to retrieve reasoning paths over wikipedia graph for
  question answering.
\newblock \emph{ICLR 2020}.

\bibitem[{Berant et~al.(2013)Berant, Chou, Frostig, and
  Liang}]{berant-etal-2013-semantic}
Jonathan Berant, Andrew Chou, Roy Frostig, and Percy Liang. 2013.
\newblock \href {https://www.aclweb.org/anthology/D13-1160} {Semantic parsing
  on {F}reebase from question-answer pairs}.
\newblock In \emph{Proceedings of the 2013 Conference on Empirical Methods in
  Natural Language Processing}, pages 1533--1544, Seattle, Washington, USA.
  Association for Computational Linguistics.

\bibitem[{Chen et~al.(2017)Chen, Fisch, Weston, and Bordes}]{chen2017reading}
Danqi Chen, Adam Fisch, Jason Weston, and Antoine Bordes. 2017.
\newblock Reading wikipedia to answer open-domain questions.
\newblock In \emph{55th Annual Meeting of the Association for Computational
  Linguistics, ACL 2017}, pages 1870--1879. Association for Computational
  Linguistics (ACL).

\bibitem[{Chen et~al.(2019)Chen, Wu, Weng, Ratner, and R{\'e}}]{chen2019slice}
Vincent~S Chen, Sen Wu, Zhenzhen Weng, Alexander Ratner, and Christopher
  R{\'e}. 2019.
\newblock Slice-based learning: A programming model for residual learning in
  critical data slices.
\newblock \emph{Advances in neural information processing systems}, 32:9392.

\bibitem[{Clark and Etzioni(2016)}]{Clark_Etzioni_2016}
Peter Clark and Oren Etzioni. 2016.
\newblock \href {https://doi.org/10.1609/aimag.v37i1.2636} {My computer is an
  honor student -- but how intelligent is it? {S}tandardized tests as a measure
  of {AI}}.
\newblock \emph{AI Magazine}, 37(1):5--12.

\bibitem[{Dehghani et~al.(2017)Dehghani, Zamani, Severyn, Kamps, and
  Croft}]{dehghani2017neural}
Mostafa Dehghani, Hamed Zamani, Aliaksei Severyn, Jaap Kamps, and W~Bruce
  Croft. 2017.
\newblock Neural ranking models with weak supervision.
\newblock In \emph{Proceedings of the 40th International ACM SIGIR Conference
  on Research and Development in Information Retrieval}, pages 65--74.

\bibitem[{Devlin et~al.(2019)Devlin, Chang, Lee, and
  Toutanova}]{devlin-etal-2019-bert}
Jacob Devlin, Ming-Wei Chang, Kenton Lee, and Kristina Toutanova. 2019.
\newblock \href {https://doi.org/10.18653/v1/N19-1423} {{BERT}: Pre-training of
  deep bidirectional transformers for language understanding}.
\newblock In \emph{Proceedings of the 2019 Conference of the North {A}merican
  Chapter of the Association for Computational Linguistics: Human Language
  Technologies, Volume 1 (Long and Short Papers)}, pages 4171--4186,
  Minneapolis, Minnesota. Association for Computational Linguistics.

\bibitem[{Ferrucci et~al.(2010)Ferrucci, Brown, Chu-Carroll, Fan, Gondek,
  Kalyanpur, Lally, Murdock, Nyberg, Prager, Schlaefer, and
  Welty}]{Watson:2010}
David Ferrucci, Eric Brown, Jennifer Chu-Carroll, James Fan, David Gondek,
  Aditya~A. Kalyanpur, Adam Lally, J.~William Murdock, Eric Nyberg, John
  Prager, Nico Schlaefer, and Chris Welty. 2010.
\newblock Building {W}atson: An overview of the deep{QA} project.
\newblock \emph{AI Magazine}, 31(3):59--79.

\bibitem[{Guu et~al.(2020)Guu, Lee, Tung, Pasupat, and Chang}]{guu2020realm}
Kelvin Guu, Kenton Lee, Zora Tung, Panupong Pasupat, and Ming-Wei Chang. 2020.
\newblock {Retrieval Augmented Language Model Pre-Training}.
\newblock \emph{ICML 2020}.

\bibitem[{Hirschman et~al.(1999)Hirschman, Light, Breck, and
  Burger}]{hirschman-etal-1999-deep}
Lynette Hirschman, Marc Light, Eric Breck, and John~D. Burger. 1999.
\newblock \href {https://doi.org/10.3115/1034678.1034731} {Deep read: A reading
  comprehension system}.
\newblock In \emph{Proceedings of the 37th Annual Meeting of the Association
  for Computational Linguistics}, pages 325--332, College Park, Maryland, USA.
  Association for Computational Linguistics.

\bibitem[{Iyyer et~al.(2014)Iyyer, Boyd-Graber, Claudino, Socher, and
  Daum{\'e}~III}]{iyyer-etal-2014-neural}
Mohit Iyyer, Jordan Boyd-Graber, Leonardo Claudino, Richard Socher, and Hal
  Daum{\'e}~III. 2014.
\newblock \href {https://doi.org/10.3115/v1/D14-1070} {A neural network for
  factoid question answering over paragraphs}.
\newblock In \emph{Proceedings of the 2014 Conference on Empirical Methods in
  Natural Language Processing ({EMNLP})}, pages 633--644, Doha, Qatar.
  Association for Computational Linguistics.

\bibitem[{Johnson et~al.(2019)Johnson, Douze, and
  J{\'e}gou}]{johnson2019billion}
Jeff Johnson, Matthijs Douze, and Herv{\'e} J{\'e}gou. 2019.
\newblock Billion-scale similarity search with gpus.
\newblock \emph{IEEE Transactions on Big Data}.

\bibitem[{Joshi et~al.(2017{\natexlab{a}})Joshi, Choi, Weld, and
  Zettlemoyer}]{joshi-etal-2017-triviaqa}
Mandar Joshi, Eunsol Choi, Daniel Weld, and Luke Zettlemoyer.
  2017{\natexlab{a}}.
\newblock \href {https://doi.org/10.18653/v1/P17-1147} {{T}rivia{QA}: A large
  scale distantly supervised challenge dataset for reading comprehension}.
\newblock In \emph{Proceedings of the 55th Annual Meeting of the Association
  for Computational Linguistics (Volume 1: Long Papers)}, pages 1601--1611,
  Vancouver, Canada. Association for Computational Linguistics.

\bibitem[{Joshi et~al.(2017{\natexlab{b}})Joshi, Choi, Weld, and
  Zettlemoyer}]{joshi2017triviaqa}
Mandar Joshi, Eunsol Choi, Daniel~S Weld, and Luke Zettlemoyer.
  2017{\natexlab{b}}.
\newblock {TriviaQA: A Large Scale Distantly Supervised Challenge Dataset for
  Reading Comprehension}.
\newblock In \emph{Proceedings of the 55th Annual Meeting of the Association
  for Computational Linguistics (Volume 1: Long Papers)}, pages 1601--1611.

\bibitem[{Karpukhin et~al.(2020)Karpukhin, Oguz, Min, Lewis, Wu, Edunov, Chen,
  and Yih}]{karpukhin2020dense}
Vladimir Karpukhin, Barlas Oguz, Sewon Min, Patrick Lewis, Ledell Wu, Sergey
  Edunov, Danqi Chen, and Wen-tau Yih. 2020.
\newblock {Dense Passage Retrieval for Open-Domain Question Answering}.
\newblock In \emph{Proceedings of the 2020 Conference on Empirical Methods in
  Natural Language Processing (EMNLP)}, pages 6769--6781.

\bibitem[{Khattab and Zaharia(2020)}]{khattab2020colbert}
Omar Khattab and Matei Zaharia. 2020.
\newblock \href {https://doi.org/10.1145/3397271.3401075} {{ColBERT: Efficient
  and Effective Passage Search via Contextualized Late Interaction over BERT}}.
\newblock In \emph{Proceedings of the 43rd International ACM SIGIR Conference
  on Research and Development in Information Retrieval}, pages 39--48.

\bibitem[{Kratzwald and
  Feuerriegel(2018)}]{kratzwald-feuerriegel-2018-adaptive}
Bernhard Kratzwald and Stefan Feuerriegel. 2018.
\newblock \href {https://doi.org/10.18653/v1/D18-1055} {Adaptive document
  retrieval for deep question answering}.
\newblock In \emph{Proceedings of the 2018 Conference on Empirical Methods in
  Natural Language Processing}, pages 576--581, Brussels, Belgium. Association
  for Computational Linguistics.

\bibitem[{Kushman et~al.(2014)Kushman, Artzi, Zettlemoyer, and
  Barzilay}]{kushman-etal-2014-learning}
Nate Kushman, Yoav Artzi, Luke Zettlemoyer, and Regina Barzilay. 2014.
\newblock \href {https://doi.org/10.3115/v1/P14-1026} {Learning to
  automatically solve algebra word problems}.
\newblock In \emph{Proceedings of the 52nd Annual Meeting of the Association
  for Computational Linguistics (Volume 1: Long Papers)}, pages 271--281,
  Baltimore, Maryland. Association for Computational Linguistics.

\bibitem[{Kwiatkowski et~al.(2019)Kwiatkowski, Palomaki, Redfield, Collins,
  Parikh, Alberti, Epstein, Polosukhin, Devlin, Lee, Toutanova, Jones, Kelcey,
  Chang, Dai, Uszkoreit, Le, and Petrov}]{kwiatkowski-etal-2019-natural}
Tom Kwiatkowski, Jennimaria Palomaki, Olivia Redfield, Michael Collins, Ankur
  Parikh, Chris Alberti, Danielle Epstein, Illia Polosukhin, Jacob Devlin,
  Kenton Lee, Kristina Toutanova, Llion Jones, Matthew Kelcey, Ming-Wei Chang,
  Andrew~M. Dai, Jakob Uszkoreit, Quoc Le, and Slav Petrov. 2019.
\newblock \href {https://doi.org/10.1162/tacl_a_00276} {Natural questions: A
  benchmark for question answering research}.
\newblock \emph{Transactions of the Association for Computational Linguistics},
  7:453--466.

\bibitem[{Lee et~al.(2019)Lee, Chang, and Toutanova}]{lee2019latent}
Kenton Lee, Ming-Wei Chang, and Kristina Toutanova. 2019.
\newblock Latent retrieval for weakly supervised open domain question
  answering.
\newblock In \emph{Proceedings of the 57th Annual Meeting of the Association
  for Computational Linguistics}, pages 6086--6096.

\bibitem[{Lewis et~al.(2019)Lewis, Liu, Goyal, Ghazvininejad, Mohamed, Levy,
  Stoyanov, and Zettlemoyer}]{lewis2019bart}
Mike Lewis, Yinhan Liu, Naman Goyal, Marjan Ghazvininejad, Abdelrahman Mohamed,
  Omer Levy, Ves Stoyanov, and Luke Zettlemoyer. 2019.
\newblock {BART}: Denoising sequence-to-sequence pre-training for natural
  language generation, translation, and comprehension.
\newblock ArXiv:1910.13461.

\bibitem[{Lewis et~al.(2020)Lewis, Perez, Piktus, Petroni, Karpukhin, Goyal,
  K{\"u}ttler, Lewis, Yih, Rockt{\"a}schel et~al.}]{lewis2020retrieval}
Patrick Lewis, Ethan Perez, Aleksandara Piktus, Fabio Petroni, Vladimir
  Karpukhin, Naman Goyal, Heinrich K{\"u}ttler, Mike Lewis, Wen-tau Yih, Tim
  Rockt{\"a}schel, et~al. 2020.
\newblock Retrieval-augmented generation for knowledge-intensive {NLP} tasks.
\newblock \emph{NeurIPS 2020}.

\bibitem[{MacAvaney et~al.(2019)MacAvaney, Yates, Hui, and
  Frieder}]{macavaney2019content}
Sean MacAvaney, Andrew Yates, Kai Hui, and Ophir Frieder. 2019.
\newblock {Content-Based Weak Supervision for Ad-Hoc Re-Ranking}.
\newblock In \emph{Proceedings of the 42nd International ACM SIGIR Conference
  on Research and Development in Information Retrieval}, pages 993--996.

\bibitem[{Min et~al.(2019{\natexlab{a}})Min, Chen, Hajishirzi, and
  Zettlemoyer}]{min2019discrete}
Sewon Min, Danqi Chen, Hannaneh Hajishirzi, and Luke Zettlemoyer.
  2019{\natexlab{a}}.
\newblock A discrete hard em approach for weakly supervised question answering.
\newblock \emph{arXiv preprint arXiv:1909.04849}.

\bibitem[{Min et~al.(2019{\natexlab{b}})Min, Chen, Zettlemoyer, and
  Hajishirzi}]{min2019knowledge}
Sewon Min, Danqi Chen, Luke Zettlemoyer, and Hannaneh Hajishirzi.
  2019{\natexlab{b}}.
\newblock Knowledge guided text retrieval and reading for open domain question
  answering.
\newblock \emph{arXiv preprint arXiv:1911.03868}.

\bibitem[{Nguyen et~al.(2016)Nguyen, Rosenberg, Song, Gao, Tiwary, Majumder,
  and Deng}]{nguyen2016ms}
Tri Nguyen, Mir Rosenberg, Xia Song, Jianfeng Gao, Saurabh Tiwary, Rangan
  Majumder, and Li~Deng. 2016.
\newblock Ms marco: A human generated machine reading comprehension dataset.
\newblock In \emph{CoCo@ NIPS}.

\bibitem[{Rajpurkar et~al.(2016{\natexlab{a}})Rajpurkar, Zhang, Lopyrev, and
  Liang}]{rajpurkar-etal-2016-squad}
Pranav Rajpurkar, Jian Zhang, Konstantin Lopyrev, and Percy Liang.
  2016{\natexlab{a}}.
\newblock \href {https://doi.org/10.18653/v1/D16-1264} {{SQ}u{AD}: 100,000+
  questions for machine comprehension of text}.
\newblock In \emph{Proceedings of the 2016 Conference on Empirical Methods in
  Natural Language Processing}, pages 2383--2392, Austin, Texas. Association
  for Computational Linguistics.

\bibitem[{Rajpurkar et~al.(2016{\natexlab{b}})Rajpurkar, Zhang, Lopyrev, and
  Liang}]{rajpurkar2016squad}
Pranav Rajpurkar, Jian Zhang, Konstantin Lopyrev, and Percy Liang.
  2016{\natexlab{b}}.
\newblock Squad: 100,000+ questions for machine comprehension of text.
\newblock In \emph{Proceedings of the 2016 Conference on Empirical Methods in
  Natural Language Processing}, pages 2383--2392.

\bibitem[{Richardson et~al.(2013)Richardson, Burges, and
  Renshaw}]{richardson-etal-2013-mctest}
Matthew Richardson, Christopher~J.C. Burges, and Erin Renshaw. 2013.
\newblock \href {https://www.aclweb.org/anthology/D13-1020} {{MCT}est: A
  challenge dataset for the open-domain machine comprehension of text}.
\newblock In \emph{Proceedings of the 2013 Conference on Empirical Methods in
  Natural Language Processing}, pages 193--203, Seattle, Washington, USA.
  Association for Computational Linguistics.

\bibitem[{Robertson and Zaragoza(2009)}]{robertson2009probabilistic}
Stephen Robertson and Hugo Zaragoza. 2009.
\newblock \emph{The probabilistic relevance framework: {BM25} and beyond}.
\newblock Now Publishers Inc.

\bibitem[{Robertson et~al.(1995)Robertson, Walker, Jones, Hancock-Beaulieu,
  Gatford et~al.}]{robertson1995okapi}
Stephen~E Robertson, Steve Walker, Susan Jones, Micheline~M Hancock-Beaulieu,
  Mike Gatford, et~al. 1995.
\newblock Okapi at {TREC-3}.
\newblock \emph{NIST Special Publication}.

\bibitem[{Vaswani et~al.(2017)Vaswani, Shazeer, Parmar, Uszkoreit, Jones,
  Gomez, Kaiser, and Polosukhin}]{Vaswani-etal:2017}
Ashish Vaswani, Noam Shazeer, Niki Parmar, Jakob Uszkoreit, Llion Jones,
  Aidan~N Gomez, \L~ukasz Kaiser, and Illia Polosukhin. 2017.
\newblock \href
  {http://papers.nips.cc/paper/7181-attention-is-all-you-need.pdf} {Attention
  is all you need}.
\newblock In I.~Guyon, U.~V. Luxburg, S.~Bengio, H.~Wallach, R.~Fergus,
  S.~Vishwanathan, and R.~Garnett, editors, \emph{Advances in Neural
  Information Processing Systems 30}, pages 5998--6008. Curran Associates, Inc.

\bibitem[{Voorhees and Tice(2000)}]{Voorhees:Tice:2000}
Ellen~M. Voorhees and Dawn~M. Tice. 2000.
\newblock \href {https://doi.org/10.1145/345508.345577} {Building a question
  answering test collection}.
\newblock In \emph{Proceedings of the 23rd Annual International ACM SIGIR
  Conference on Research and Development in Information Retrieval}, pages
  200--207, New York, NY, USA. Association for Computing Machinery.

\bibitem[{Wang et~al.(2019)Wang, Ng, Ma, Nallapati, and Xiang}]{wang2019multi}
Zhiguo Wang, Patrick Ng, Xiaofei Ma, Ramesh Nallapati, and Bing Xiang. 2019.
\newblock {Multi-passage BERT: A Globally Normalized BERT Model for Open-domain
  Question Answering}.
\newblock In \emph{Proceedings of the 2019 Conference on Empirical Methods in
  Natural Language Processing and the 9th International Joint Conference on
  Natural Language Processing (EMNLP-IJCNLP)}, pages 5881--5885.

\bibitem[{Wolf et~al.(2020)Wolf, Debut, Sanh, Chaumond, Delangue, Moi, Cistac,
  Rault, Louf, Funtowicz, Davison, Shleifer, von Platen, Ma, Jernite, Plu, Xu,
  Le~Scao, Gugger, Drame, Lhoest, and Rush}]{wolf-etal-2020-transformers}
Thomas Wolf, Lysandre Debut, Victor Sanh, Julien Chaumond, Clement Delangue,
  Anthony Moi, Pierric Cistac, Tim Rault, Remi Louf, Morgan Funtowicz, Joe
  Davison, Sam Shleifer, Patrick von Platen, Clara Ma, Yacine Jernite, Julien
  Plu, Canwen Xu, Teven Le~Scao, Sylvain Gugger, Mariama Drame, Quentin Lhoest,
  and Alexander Rush. 2020.
\newblock \href {https://doi.org/10.18653/v1/2020.emnlp-demos.6} {Transformers:
  State-of-the-art natural language processing}.
\newblock In \emph{Proceedings of the 2020 Conference on Empirical Methods in
  Natural Language Processing: System Demonstrations}, pages 38--45, Online.
  Association for Computational Linguistics.

\bibitem[{Yang et~al.(2018)Yang, Fang, and Lin}]{yang2018anserini}
Peilin Yang, Hui Fang, and Jimmy Lin. 2018.
\newblock Anserini: Reproducible ranking baselines using lucene.
\newblock \emph{Journal of Data and Information Quality (JDIQ)}, 10(4):1--20.

\bibitem[{Yang et~al.(2019)Yang, Xie, Lin, Li, Tan, Xiong, Li, and
  Lin}]{yang2019end}
Wei Yang, Yuqing Xie, Aileen Lin, Xingyu Li, Luchen Tan, Kun Xiong, Ming Li,
  and Jimmy Lin. 2019.
\newblock End-to-end open-domain question answering with bertserini.
\newblock In \emph{Proceedings of the 2019 Conference of the North American
  Chapter of the Association for Computational Linguistics (Demonstrations)},
  pages 72--77.

\bibitem[{Yang et~al.(2015)Yang, Yih, and Meek}]{yang-etal-2015-wikiqa}
Yi~Yang, Wen-tau Yih, and Christopher Meek. 2015.
\newblock \href {https://doi.org/10.18653/v1/D15-1237} {{W}iki{QA}: A challenge
  dataset for open-domain question answering}.
\newblock In \emph{Proceedings of the 2015 Conference on Empirical Methods in
  Natural Language Processing}, pages 2013--2018, Lisbon, Portugal. Association
  for Computational Linguistics.

\bibitem[{Zhang et~al.(2020)Zhang, Xiong, Liu, and Liu}]{zhang2020selective}
Kaitao Zhang, Chenyan Xiong, Zhenghao Liu, and Zhiyuan Liu. 2020.
\newblock {Selective Weak Supervision for Neural Information Retrieval}.
\newblock In \emph{Proceedings of The Web Conference 2020}, pages 474--485.

\end{thebibliography}
\bibliographystyle{acl_natbib}

\end{document}